\documentclass[11pt,twocolumn,twoside]{article}
\usepackage{iccr2024}
\usepackage{multirow}

\begin{document}

\title{An anatomically-informed correspondence initialisation method to improve learning-based registration for radiotherapy} 

\author[1]{Edward~G.~A.~Henderson}
\author[1]{Marcel~van~Herk}
\author[2]{Andrew~F.~Green}
\author[1]{Eliana~M.~Vasquez~Osorio}

\affil[1]{Division of Cancer Sciences, The University of Manchester, M13 9PL, UK}

\affil[2]{European Bioinformatics Institute, EMBL-EBI, Cambridge, UK}

\maketitle
\thispagestyle{fancy}


\begin{customabstract}
We propose an anatomically-informed initialisation method for interpatient CT non-rigid registration (NRR), using a learning-based model to estimate correspondences between organ structures. A thin plate spline (TPS) deformation, set up using the correspondence predictions, is used to initialise the scans before a second NRR step. We compare two established NRR methods for the second step: a B-spline iterative optimisation-based algorithm and a deep learning-based approach. Registration performance is evaluated with and without the initialisation by assessing the similarity of propagated structures. Our proposed initialisation improved the registration performance of the learning-based method to more closely match the traditional iterative algorithm, with the mean distance-to-agreement reduced by 1.8mm for structures included in the TPS and 0.6mm for structures not included, while maintaining a substantial speed advantage (5 vs. 72 seconds).
\end{customabstract}

\section{Introduction}
Non-rigid registration (NRR) is widely used for many applications in radiotherapy to spatially align images, including fusion of complementary images, contour propagation, evaluating treatment response, dose mapping~\cite{Brock2017,Murr2023}. Traditional image registration algorithms function by performing iterative updates to parameters of a chosen deformation model to optimise either an intensity- and/or feature-based metric~\cite{Brock2017}. Intensity-based metrics use the images greyscale intensity information directly to evaluate similarity. Feature-based metrics use features extracted from the images such as anatomical landmarks, structure boundaries, contours and fiducial markers to evaluate alignment.

Hybrid image registration techniques have been developed which aim to combine the strengths of intensity- and feature-based algorithms, leveraging greyscale information as well as anatomical structures or landmarks to guide the registration~\cite{Weistrand2015, Kim2013}. However, hybrid methods tend to be slow and labour intensive, taking a long time to set up as they require landmark annotation or structure contouring.

Recently image registration methods using deep learning (DL) have garnered much attention as they can offer large time savings over iterative algorithms~\cite{Zou2022}. DL approaches are trained ahead of time on large datasets of examples; once trained, these methods can estimate a full deformation vector field in a single function evaluation, often taking less than a second on modern hardware~\cite{Teuwen2022}. However, current learning-based approaches are predominantly applied to easier tasks such as the registration of brain MRI scans where the inter-subject variability is relatively low and the soft-tissue contrast is high~\cite{Zou2022,Heinrich2022}. There has been little evidence to show that learning-based registration methods are suitable for large deformations, such as inter-subject CT registration~\cite{Bhattacharjee2021,Teuwen2022}.

Many learning-based registration approaches are trained to optimise intensity-based metrics. Some can leverage segmentations present in the images, but only at training time through semi-supervised learning~\cite{Zou2022}. In this study we incorporate ideas from hybrid registration methods to improve learning-based registration performance. To achieve this we introduce a new anatomical correspondence-informed non-rigid initialisation to align labelled anatomical structures prior to using established methods to register non-labelled tissue.

\section{Materials and Methods}
We used an open-access dataset of 31 head and neck CT scans with highly consistent organ-at-risk (OAR) segmentations~\cite{nikolov2018}.
\vspace{-1mm}
\subsection{Registration pipelines}
We propose the use of a previously developed learning-based correspondence model~\cite{Henderson2023} and a thin plate spline (TPS) to form an anatomically informed method to initialise inter-patient non-rigid registration. The impact of the proposed initialisation was compared for two established NRR methods: a) NiftyReg, a traditional B-spline iterative-optimisation based image registration method~\cite{Modat2010} and b) Voxelmorph, an unsupervised, DL-based approach~\cite{Balakrishnan2019}. From now on, the proposed initialisation method is referred to as CorrTPS.
\vspace{-1mm}
\subsubsection{Anatomically informed initialisation (\textit{CorrTPS})}
\label{ssec:CorrTPS}
\paragraph{Correspondence model}
In a previous study, we showed an unsupervised geometric learning model could be trained to identify dense, point-to-point, correspondences for head and neck OARs~\cite{Henderson2023}.
The developed model can automatically identify corresponding anatomical points on the surface of 3D organ shapes and was shown to be effective for the brainstem, mandible, parotid glands, spinal cord and submandibular glands. The input for the correspondence model is two organ shapes which are converted to triangular meshes. The model predicts an assignment linking all vertices from the source to the target organ. Correspondence predictions are based on the organ geometry alone, however, CT imaging was utilised in the training process to improve prediction quality~\cite{Henderson2023}.

\begin{figure*}[t]
\centerline{\includegraphics[width=18.5cm]{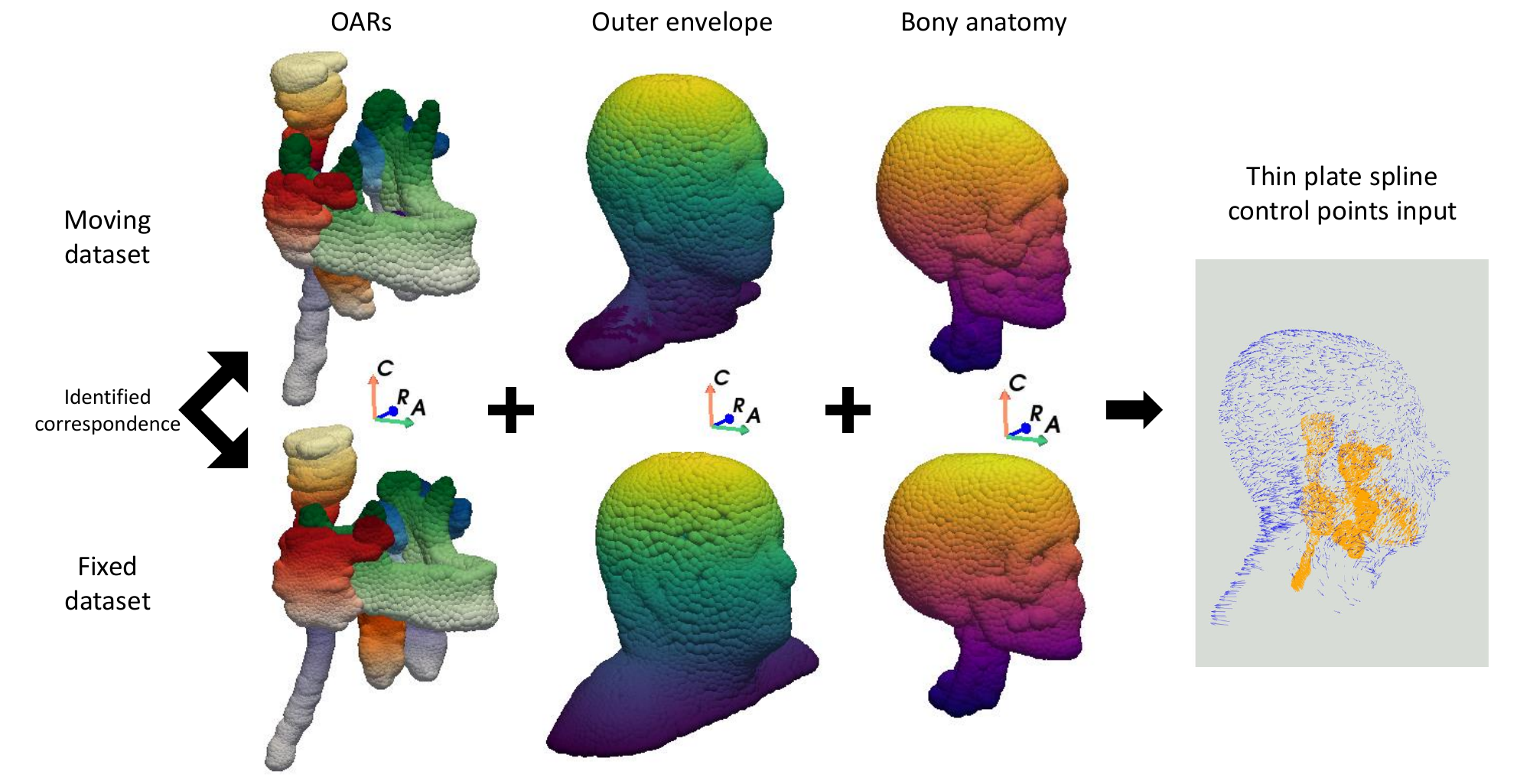}}
\caption{Correspondence of each structure between the fixed and moving datasets, represented by matching colours. Different colour maps are used for each structure and identified correspondences are only shown in the cranio-caudal direction for clarity.  The resulting deformation vectors (OARs in orange, bony anatomy and envelope in blue) are the input for the TPS to set up the non-rigid initialisation.
} \label{reg_correspondences_fig}
\end{figure*}

\paragraph{Mesh generation}
To create triangular meshes of the OAR structures, we used the marching cubes algorithm followed by Taubin smoothing, mesh decimation, optimisation and simplification operations. The full details of this process have been described fully in our previous work~\cite{Henderson2023}.

A segmentation of the bony anatomy was acquired automatically from the CT scan by removing the table, thresholding at 400 Hounsfield units (HU), cropping at the base of the neck, and meshing using the marching cubes algorithm. A mesh of the external envelope comprising of the patient's body was generated using a lower threshold level of -200 HU.

\paragraph{Correspondence inference}
For non-rigid initialisation, dense correspondences were estimated for all segmented structures in the HN, as well as the bony anatomy and outer envelope of each patient (Figure~\ref{reg_correspondences_fig}).

\paragraph{Thin plate spline setup}
Once estimated, the correspondences were used as the control points to set up a TPS deformation model~\cite{Bookstein1989}. The TPS is generalizable to 3D and has a regularisation parameter, $\lambda$\textsubscript{TPS}, which controls the allowed degree of local deformations.

Control points from each structure were gathered and a single TPS was created for each registration pair. In total, the TPS deformations were performed using about 14,000 control points (range: 10,500 - 16,000). Regions that were identified as null correspondences were excluded from the control point set driving the TPS. In this way, we prevent the creation of anatomically infeasible scenarios, e.g. inverting and folding around tissue mismatches.

For each fixed-moving image pair, we generated a displacement vector field by evaluating the TPS on the grid positions of the moving image. Then, we resampled the fixed image by inverse warping using the \textit{DisplacementFieldTransform} from the \textit{SimpleITK} package. We refer to these as non-rigidly initialised scans. A second NRR algorithm was then used to fine-tune the registration using the image intensities for the whole anatomy.

\subsubsection{Non-rigid registration}
To demonstrate the added value of CorrTPS, we compared the performance of two established intensity-based NRR methods with and without our proposed initialisation.

\begin{table*}[ht]
\caption{Mean execution time for all registration pipelines. These times include all necessary pre-processing steps for each method such as rigid pre-alignment, cropping, normalisation, and execution of the second non-rigid registration. All pipelines required rigid pre-alignment. The average time added by CorrTPS is shown in brackets and separated by the contributions of mesh generation and optimisation, correspondence estimation, TPS fitting and resampling respectively.}
\label{reg_execution_times}
\small
\centering
\begin{tabular}{cccc}
    & \begin{tabular}{@{}c@{}}Rigid pre-\\alignment (s)\end{tabular} & \begin{tabular}{@{}c@{}}Pre-processing steps and\\non-rigid registration (s)\end{tabular} & \begin{tabular}{@{}c@{}}Overall time to register\\image pair (s)\end{tabular} \\ \hline
Voxelmorph & \multirow{4}{*}{1.6} & 0.3 & 1.9 \\
Voxelmorph + CorrTPS &  & (1.2 + 0.1 + 1.4 + 0.2) + 0.3 & 4.8 \\
NiftyReg &  & 70.7 & 72.3 \\
NiftyReg + CorrTPS &  & (1.2 + 0.1 + 1.4 + 0.2) + 70.7 & 75.2
\end{tabular}
\end{table*}

\normalsize
\paragraph{Deep learning-based registration (Voxelmorph)}
We trained multiple Voxelmorph models from scratch~\cite{Balakrishnan2019}. Image volumes were automatically cropped to dimensions of $128\times288\times480$ such that the whole HN anatomy was within the field-of-view while retaining the original voxel spacing. Additionally, the images were normalised from Hounsfield units onto the range [0,1] using contrast windowing with wide window settings (W~1600HU, L~0HU). The weight of the smoothing loss term which regularises the predicted deformations was set to 0.005.

When performing the five-fold cross-validation, each fold contains 7 testing patients. This leads to a $7\times6$ testing pairs for each fold (trivial self-registration cases excluded). The same Voxelmorph models were applied to the non-rigidly initialised scans.

\paragraph{Iterative registration (NiftyReg)}
NiftyReg was first used to non-rigidly register the CT scans after rigid pre-alignment alone~\cite{Modat2010}. Following an empirical parameter evaluation, the configuration settled upon was a GPU-accelerated cubic B-spline using normalised mutual information loss with a bending energy of 0.001. The same configuration of NiftyReg was then used to register the non-rigidly initialised CT scans.
\vspace{-6mm}
\subsection{Evaluation}
\vspace{-1mm}
For every pipeline, we report the execution time including all the necessary pre-processing steps and the results image- and segmentation-based metrics.

The mean distance-to-agreement (mDTA) was calculated for each of the structures included within CorrTPS as well as the brain, cochleae, optic nerves, orbits following non-rigid alignment of the moving anatomy to the fixed reference patient. Wilcoxon signed-rank hypothesis tests were used to compare each registration algorithm with and without the CorrTPS initialisation. A Bonferroni correction was applied to minimise the risk of type I errors from multiple hypothesis tests (corrected p value: 0.0045).
\vspace{-3mm}
\paragraph{Implementation details}
Model training and evaluation was performed on a workstation with an AMD Ryzen 9 3950X 16-Core Processor, 64GB of RAM and an NVidia GeForce RTX 4090 GPU.
\vspace{-3mm}
\section{Results}
\vspace{-3mm}
The average runtime for each approach is shown in Table~\ref{reg_execution_times}.

\begin{figure}[t]
\centerline{\includegraphics[width=\columnwidth]{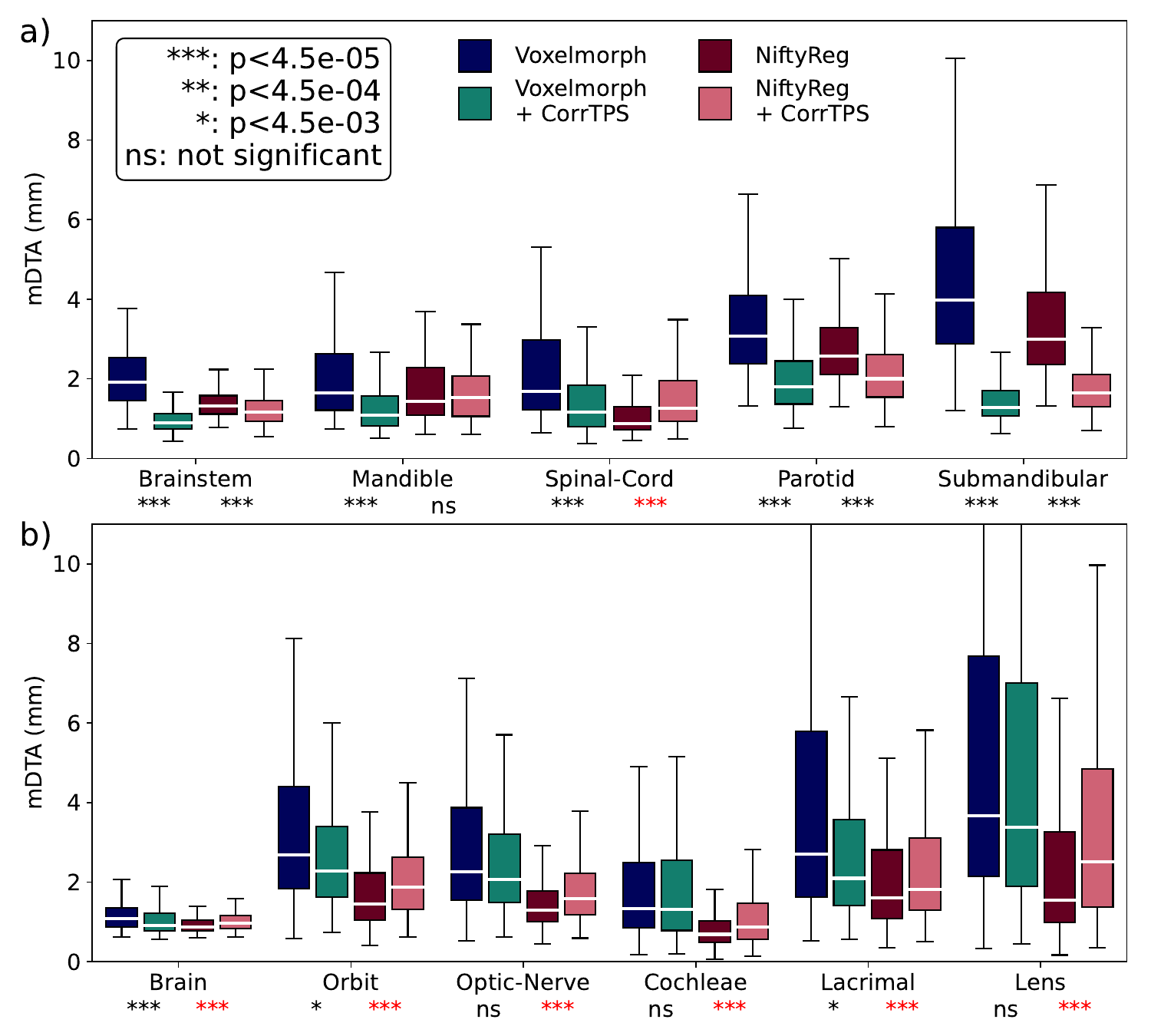}}
\caption{Mean distance-to-agreement results for each registration pipeline. The structures in a) were included in CorrTPS, i.e., their surface correspondence was used to set up the TPS, whereas the structures in b) were not. Significant deviations between methods with and without CorrTPS, as identified with a Wilcoxon signed-rank test, are indicated. Black asterisks show an improvement using CorrTPS, whereas the red show performance reductions.} \label{reg_seg_results}
\end{figure}

Figure~\ref{reg_seg_results}a shows boxplots of the mDTA results for the organ structures included in the CorrTPS, while \ref{reg_seg_results}b shows similar results for structures not included in CorrTPS. Significant improvements in registration performance were observed for most structures included in CorrTPS across both methods when using the proposed initialisation. For structures not included in CorrTPS, some improvements were observed for the learning-based registration method (the brain, orbits and lacrimal glands), however, the performance of the traditional registration algorithm was degraded.

\vspace{-2mm}
\section{Discussion}
\vspace{-2mm}
\label{reg_discussion}
We proposed a novel method to non-rigidly pre-register an image pair prior to inter-subject registration of head and neck CT scans using correspondence of anatomical structures. The initialisation method is based on a geometric learning-based correspondence model which, when coupled with a TPS deformation, allowed the moving scans to be deformed to align segmented OARs before a final registration using established NRR methods. Our initialisation approach showed promise to improve the registration performance of the learning-based NRR method, Voxelmorph, improving the quality of the registrations in terms of the segmentation-based evaluation measure. The performance of our proposed method with the traditional iterative algorithm, NiftyReg, was less clearly defined, with some improvements for structures included in the initialisation and degradations for those not included. The use of CorrTPS required a limited increase in execution time.

In terms of the mDTA, the benefit of our proposed initialisation method is evident for structures included within CorrTPS. The one exception to this was for the spinal cord when using NiftyReg, in which the use of the initialisation method harmed registration performance.

Little improvement was observed when using our method for the structures not included in CorrTPS, which were predominantly in the superior region of the head. This is likely due to the TPS deformation in this region comprising of an interpolation between the envelope and bony anatomy and the controlling structures which were mainly further inferior in the head and neck region. This shows a sensitivity of the proposed method to the spatial distribution of controlling structures. For registrations with the iterative algorithm, the use of the proposed method degraded the performance for all structures not included in CorrTPS.

While we included some of the structures in this study and excluded others from CorrTPS, the proposed method is applicable to any of the OAR structures segmented for this study, i.e., the complete set of structures could also be included in CorrTPS, thereby fitting a wider distribution of control points within the head and neck anatomy. The implementation of TPS used is fast, estimating a deformation within a second for $\sim14,000$ control points.

We deployed our proposed initialisation method with an intensity-based traditional NRR algorithm and a DL-based method. In future it would be useful to compare the performance of our hybrid-inspired registration framework against a hybrid registration methods such as ANACONDA~\cite{Weistrand2015} to further assess the efficacy of the proposed method.

In this study we focused solely on performing inter-patient registration, despite the task of intra-patient registration being more common within the radiotherapy treatment pathway. Inter-patient registration is important for tasks such as outcome modelling and can be challenging due to the increased variability in anatomy between subjects. We hypothesize that our initialisation method would be particularly effective in intra-patient cases such as in the re-irradiation setting, where large and complex anatomical changes are common, e.g. resulting from surgical interventions, which hinder the accuracy of currently available image registration algorithms~\cite{Murr2023, Nenoff2023}.

The segmentations used in this study were manually performed. Future introduction of auto-segmentation could mitigate the impact of inter- and intra-observer variation and improve correspondence reliability. Additionally, the use of auto-segmentations with the proposed framework would facilitate the implementation of a fully automated anatomically-informed registration pipeline.

\section{Conclusion}
The proposed method to initialise an image pair prior to NRR using anatomical correspondence shows particular promise to enhance the registration performance of DL-based registration methods to more closely match that of a traditional iterative NRR algorithm, whilst maintaining a substantial speed advantage. With further development and validation, the proposed approach could additionally raise the performance of traditional iterative NRR algorithms.

\printbibliography

@article{Balakrishnan2019,
  doi = {10.1109/tmi.2019.2897538},
  year = {2019},
  month = {8},
  publisher = {Institute of Electrical and Electronics Engineers ({IEEE})},
  volume = {38},
  number = {8},
  pages = {1788--1800},
  author = {Guha Balakrishnan and Amy Zhao and Mert R. Sabuncu and John Guttag and Adrian V. Dalca},
  title = {{VoxelMorph}: A Learning Framework for Deformable Medical Image Registration},
  journal = {{IEEE} Transactions on Medical Imaging}
}

@article{Bhattacharjee2021,
  doi = {10.1016/j.irbm.2020.04.002},
  year = {2021},
  month = {4},
  publisher = {Elsevier {BV}},
  volume = {42},
  number = {2},
  pages = {94--105},
  author = {R. Bhattacharjee and F. Heitz and V. Noblet and S. Sharma and N. Sharma},
  title = {Evaluation of a Learning-based Deformable Registration Method on Abdominal {CT} Images},
  journal = {{IRBM}}
}

@article{Bookstein1989,
  doi = {10.1109/34.24792},
  year = {1989},
  month = {6},
  publisher = {Institute of Electrical and Electronics Engineers ({IEEE})},
  volume = {11},
  number = {6},
  pages = {567--585},
  author = {F.L. Bookstein},
  title = {Principal warps: thin-plate splines and the decomposition of deformations},
  journal = {{IEEE} Transactions on Pattern Analysis and Machine Intelligence}
}

@article{Brock2017,
  doi = {10.1002/mp.12256},
  year = {2017},
  month = {5},
  publisher = {Wiley},
  volume = {44},
  number = {7},
  pages = {e43--e76},
  author = {Kristy K. Brock and Sasa Mutic and Todd R. McNutt and Hua Li and Marc L. Kessler},
  title = {Use of image registration and fusion algorithms and techniques in radiotherapy: Report of the {AAPM} Radiation Therapy Committee Task Group No. 132},
  journal = {Medical Physics}
}

@incollection{Heinrich2022,
  doi = {10.1007/978-3-031-11203-4_10},
  year = {2022},
  publisher = {Springer International Publishing},
  pages = {85--95},
  author = {Mattias P. Heinrich and Lasse Hansen},
  title = {Voxelmorph++},
  booktitle = {{Biomedical Image Registration}}
}

@incollection{Henderson2023,
  doi = {10.1007/978-3-031-46914-5_7},
  year = {2023},
  publisher = {Springer Nature Switzerland},
  pages = {75--89},
  author = {Edward G. A. Henderson and Marcel {van Herk} and Andrew F. Green and Eliana M. {Vasquez Osorio}},
  title = {{Unsupervised Correspondence with Combined Geometric Learning and Imaging for Radiotherapy Applications}},
  booktitle = {Shape in Medical Imaging}
}

@article{Kim2013,
  doi = {10.1088/0031-9155/58/22/8077},
  year = {2013},
  month = {10},
  publisher = {{IOP} Publishing},
  volume = {58},
  number = {22},
  author = {Jinkoo Kim and Sanath Kumar and Chang Liu and Hualiang Zhong and Deepak Pradhan and Mira Shah and Richard Cattaneo and Raphael Yechieli and Jared R Robbins and Mohamed A Elshaikh and Indrin J Chetty},
  title = {A novel approach for establishing benchmark {CBCT}/{CT} deformable image registrations in prostate cancer radiotherapy},
  journal = {Physics in Medicine \& Biology}
}

@article{Modat2010,
  doi = {10.1016/j.cmpb.2009.09.002},
  year = {2010},
  month = jun,
  publisher = {Elsevier {BV}},
  volume = {98},
  number = {3},
  pages = {278--284},
  author = {Marc Modat and Gerard R. Ridgway and Zeike A. Taylor and Manja Lehmann and Josephine Barnes and David J. Hawkes and Nick C. Fox and S{\'{e}}bastien Ourselin},
  title = {Fast free-form deformation using graphics processing units},
  journal = {Computer Methods and Programs in Biomedicine}
}

@article{Murr2023,
  doi = {10.1016/j.radonc.2023.109527},
  year = {2023},
  month = {5},
  publisher = {Elsevier {BV}},
  volume = {182},
  pages = {109527},
  author = {Martina Murr and Kristy K. Brock and Marco Fusella and Nicholas Hardcastle and Mohammad Hussein and Michael G. Jameson and Isak Wahlstedt and Johnson Yuen and Jamie R. McClelland and Eliana M. {Vasquez Osorio}},
  title = {Applicability and usage of dose mapping/accumulation in radiotherapy},
  journal = {Radiotherapy and Oncology}
}

@article{Nenoff2023,
  title = {Review and recommendations on deformable image registration uncertainties for radiotherapy applications},
  DOI = {10.1088/1361-6560/ad0d8a},
  journal = {Physics in Medicine \& Biology},
  publisher = {IOP Publishing},
  author = {Nenoff,  Lena and Amstutz,  Florian and Murr,  Martina and Archibald-Heeren,  Ben and Fusella,  Marco and Hussein,  Mohammad and Lechner,  Wolfgang and Zhang,  Ye and Sharp,  Gregory C and {Vasquez Osorio},  Eliana M},
  year = {2023},
  month = {11} 
}

@article{nikolov2018,
  author = {Stanislav Nikolov and Sam Blackwell and Ruheena Mendes and Jeffrey De Fauw and Clemens Meyer and Cían Hughes and Harry Askham and Bernardino Romera-Paredes and Alan Karthikesalingam and Carlton Chu and Dawn Carnell and Cheng Boon and Derek D'Souza and Syed Ali Moinuddin and Kevin Sullivan and DeepMind Radiographer Consortium and Hugh Montgomery and Geraint Rees and Ricky Sharma and Mustafa Suleyman and Trevor Back and Joseph R. Ledsam and Olaf Ronneberger},
  title = {Deep learning to achieve clinically applicable segmentation of head and neck anatomy for radiotherapy},
  journal = {ArXiv e-prints},
  year = {2018},
  doi = {10.48550/arXiv.1809.04430}
}

@article{Teuwen2022,
  doi = {10.1016/j.semradonc.2022.06.003},
  year = {2022},
  month = {10},
  publisher = {Elsevier {BV}},
  volume = {32},
  number = {4},
  pages = {330--342},
  author = {Jonas Teuwen and Zeno A.R. Gouw and Jan-Jakob Sonke},
  title = {Artificial Intelligence for Image Registration in Radiation Oncology},
  journal = {Seminars in Radiation Oncology}
}

@article{Weistrand2015,
  doi = {10.1118/1.4894702},
  year = {2015},
  month = {1},
  publisher = {Wiley},
  volume = {42},
  number = {1},
  pages = {40--53},
  author = {Ola Weistrand and Stina Svensson},
  title = {The {ANACONDA} algorithm for deformable image registration in radiotherapy},
  journal = {Medical Physics}
}

@article{Zou2022,
  doi = {10.3389/fonc.2022.1047215},
  year = {2022},
  month = {12},
  publisher = {Frontiers Media {SA}},
  volume = {12},
  author = {Jing Zou and Bingchen Gao and Youyi Song and Jing Qin},
  title = {A review of deep learning-based deformable medical image registration},
  journal = {Frontiers in Oncology}
}

\end{document}